\documentclass[11pt,letterpaper]{article}
\usepackage{emnlp2017}
\usepackage{times}
\usepackage{latexsym}
\usepackage{amsmath}
\usepackage{graphicx}
\usepackage{color}
\usepackage{xspace}
\usepackage{array}
\usepackage{url}
\usepackage{natbib}
\usepackage{array,multirow}
\usepackage{booktabs}
\usepackage{tikz-dependency}

\usepackage[aboveskip=-2pt,belowskip=+3pt]{subcaption}
\definecolor{wine}{rgb}{0.65,0,0.35}

\emnlpfinalcopy

\author{
	Justine Zhang,$^{1}$ Arthur Spirling,$^{2}$ 
	Cristian Danescu-Niculescu-Mizil$^{1}$\\
        $^1$Cornell University, $^2$New York University\\
  	\small {\tt \href{mailto:jz727@cornell.edu}{jz727@cornell.edu}},
	\small {\tt \href{mailto:arthur.spirling@nyu.edu}{arthur.spirling@nyu.edu}},
	{\tt \href{mailto:cristian@cs.cornell.edu}{cristian@cs.cornell.edu}} \\
}

\newcommand{\cut}[1]{}
\definecolor{wine}{rgb}{0.7,0.1,0.2}

\newcommand{\cd}[1]{}
\newcommand{\justine}[1]{}

\newcommand{\xhdr}[1]{{\noindent\bfseries #1.}}

\title{Asking too much?  The rhetorical role of questions in political discourse
}

\begin{document}

\maketitle
\begin{abstract}
Questions play a prominent role in social interactions, performing rhetorical functions that go beyond that of simple informational exchange.  The surface form of a question can signal the intention and background of the person asking it, as well as the nature of their relation with the interlocutor.  While the informational nature of questions has been extensively
examined in the context of question-answering applications, their rhetorical aspects have been largely understudied. 

In this work we introduce an unsupervised methodology for extracting surface motifs that recur in questions, and for grouping them according to their latent rhetorical role.  By applying this framework to the setting of question sessions 
 in the UK parliament, 
we show that the resulting typology encodes key aspects of the political discourse---such as the bifurcation in questioning behavior between government and opposition parties---and reveals new insights into the effects %
of a legislator's tenure and political career ambitions.

\end{abstract}

\section{Introduction}
\label{sec:intro}

\begin{quote}
\small
``We'd now like to open the floor to shorter \\ speeches disguised as questions...''
\end{quote}
\vspace{-0.3cm}
\begin{flushright}
\small
-- Steve Macone, New Yorker cartoon caption
\end{flushright}

Why do we ask questions? Perhaps we are seeking factual information that others hold, or maybe we are requesting a favor. Alternatively we could be simply making a rhetorical point, perhaps at the start of an academic paper. 
 
Questions play a prominent role in social interactions \cite{goffman1976replies}, performing a multitude of rhetorical functions that go beyond mere factual information gathering
 \cite{Kearsley:JournalOfPsycholinguisticResearch:}.  While the informational component of questions has been well-studied in the context of question-answering applications, there is relatively little computational work addressing the rhetorical and social role of these basic dialogic units.

One domain where  questions have a particularly salient rhetorical role is politics. 
The ability to question the actions and intentions of governments
is a crucial part of democracy \cite{Pitkin67}, particularly in parliamentary systems.
 Consequently, scholars have studied parliamentary questions in detail, in terms of their origins \cite{ChesterBowring1962}, their institutionalization \cite{EggersSpirling2012} and their importance for oversight \cite{ProkschSlapin11}. 
In particular, the United Kingdom's House of Commons, renowned for theatrical questions periods, 
has been studied in some depth. 
However, those accounts are largely qualitative in nature \cite{BullWells2012,Batesetal2014}.

\xhdr{The present work: methodology} In order to approach these problems computationally, we introduce an unsupervised framework to structure the space of questions according to their rhetorical role.  
First, we identify common ways in which questions are phrased. To this end, we automatically extract these recurring surface forms, or \textit{motifs}, based on the lexico-syntactic structure of the questions posed (Section~\ref{sec:motifs}). 
To capture rhetorical aspects we then group these motifs according to their role, %
relying on the intuition that 
this role is 
encoded in
the type of {\em answer} a question receives. %
To operationalize this intuition we construct a latent question-answer space in which question motifs triggering similar answers are mapped to the same region (Section~\ref{sec:qtypes}).

\xhdr{The present work: application} We apply this general framework to the political discourse that occurs during parliamentary question sessions in the British House of Commons, a new dataset which we make publicly available (Section~\ref{sec:data}).  %
Our framework extracts intuitive question types ranging from narrow factual queries to pointed criticisms disguised as questions (Section~\ref{sec:qtypes}, Table~\ref{tab:qtypes}). We validate our framework by aligning these types with prior understandings of parliamentary proceedings from the political science literature (Section~\ref{sec:validation}). In particular, previous work \cite{Batesetal2014} has categorized questions asked in Parliament according to the intentions of the asker (e.g., to help the answerer, or to adversarially put them on the spot); we find a clear, predictive mapping between these expert-coded categories and the induced typology. 
We further show that the types of questions specific legislators tend to ask vary with whether they are part of the governing or opposition party, consistent with well-established accounts of partisan differences
 \cite{Cowley02,SpirlingMcLean07,EggersSpirling2012}. Concretely, government legislators exhibit a preference for overtly friendly questions, while the opposition slants towards more aggressive question types. %

We then apply our methodology to provide new insights into how a legislator's questioning behavior varies with their career trajectory. The pressures faced by legislators at various stages in their career are cross-cutting, and multiple possible hypotheses emerge. Younger, more enthusiastic legislators may be motivated to ask harder-hitting questions, but risk being passed over for future promotion if they are too combative \cite{Cowley02}. Older legislators, whose opportunities for promotion are largely behind them and hence have ``less to lose'', may act more aggressively \cite{BenedettoHix07}; or simply seek a quiet path to retirement. Viewing each group's behavior through the questions they ask brings evidence for the latter hypothesis that 
more tenured
 legislators are more aggressive, even when questioning their own leaders. In this way, their presence in the House of Commons, and their refusal to simply `keep their heads down', facilitates a core component of democracy. 

\section{Related Work}
\label{sec:relwork}
	\xhdr{Question-answering}  Computationally, questions have received considerable attention in the context of question-answering (QA) systems---for a survey see \newcite{gupta2012survey}---with an emphasis on understanding their information need \cite{Harabagiu:AdvancesInOpenDomainQuestionAnswering:2008}. Techniques have been developed to categorize questions based on the nature of these information needs in the context of the TREC QA challenge \cite{harabagiu2000falcon}, and to identify questions asking for similar information \cite{shtok2012learning,Zhang:2017:DDP:3038912.3052701,Jeon:2005:FSS:1076034.1076156}; questions have also been classified by topic \cite{cao2010generalized} and quality \cite{treude2011programmers,Ravi:Icwsm:2014}. In contrast, our work is not concerned with the information need central to QA applications, and instead focuses on the rhetorical aspect of questions.

	\xhdr{Question types} To facilitate retrieval of frequently asked questions, \citet{lytinen2002use} manually developed a typology of surface question forms (e.g., `what'- and `why'-questions) starting from Lehnerts' conceptual question categories \cite{lehnert1978process}. Question types were also hand annotated for dialog-act labeling, 
	distinguishing between yes-no, wh-, open-ended and rhetorical questions \cite{dhillon2004meeting}. To complement this line of work, this paper introduces a completely unsupervised methodology to automatically build a domain-tailored question typology, bypassing the need for human annotation.

	\xhdr{Pragmatic dimensions} 
One important pragmatic dimension of questions that has been previously  studied computationally is their level of politeness \cite{Danescu-Niculescu-Mizil+al:13b,Aubakirova:ProceedingsOfEmnlp:2016}; in the specific context of making requests, politeness was shown to correlate with the social status of the asker.
\citet{Sachdeva:2017:CSC:2998181.2998292} studied another rhetorical aspect by examining linguistic attributes distinguishing serviceable requests addressed to police on social media from general conversation.
Previous research has also been directed at identifying rhetorical questions \cite{bhattasali2015automatic} and understanding the motivations of their ``askers'' \cite{ranganath2016identifying}.  
Using the relationship between questions and answers, our work examines the rhetorical and social aspect of questions without predefining a pragmatic dimension and without relying on labeled data. 
We also complement these efforts in analyzing a broader range of situations in which questions may be posed without an information-seeking intent.

	\xhdr{Political discourse} Finally, our work contributes to a rapidly growing area of NLP applications to political domains \cite[inter alia]{Monroe:PoliticalAnalysis:2008,Card:Emnlp:2016,gonzalez2010structure,niculae2015quotus,Grimmer:PoliticalAnalysis:2013,Grimmer:AmericanPoliticalScienceReview:2012,iyyerpolitical}. Particularly relevant are  applications to discourse in congressional and parliamentary settings \cite{Thomas:2006:GOV:1610075.1610122,Boydstun:ProceedingsOfTheApsa:2014,rheault2016measuring}.

\section{Data: Parliamentary Question Periods}
\label{sec:data}
The bulk of our analysis focuses on the questions asked, and responses given during parliamentary question periods in the British House of Commons. Below, we provide a brief overview of key features of this political system in general, as well as a description of the question period setting.

\xhdr{Parliamentary systems} 
 Legislators in the House of Commons ({\em Members of Parliament}, henceforth {\em MPs} or {\em members}) belong to two main voting and debating {\em affiliations}: a {\em government} party which controls the executive and holds a majority of the seats in the chamber, and a set of {\em opposition} parties.\footnote{We use {\em affiliation} to refer broadly to the government and opposition roles, independent of the identity of the current government and opposition parties. In subsequent analysis we only consider the largest, ``official'' opposition party as the opposition.}
The executive is headed by the Prime Minister (PM) and run by a cabinet of {\em ministers}, high-ranking government MPs responsible for various departments such as finance and education. 

\xhdr{Question periods}
The House of Commons holds weekly, moderated {\em question periods}, in which MPs of all affiliations take turns to ask questions to (and theoretically receive answers from) government ministers for each department regarding their specific domains. 
Such events are a primary way in which legislators
hold senior policy-makers responsible for their decisions. In practice, beyond narrow requests for information about specific policy points, MPs use their questions to critique or praise the government, or to self-promote; indeed, certain sessions, such as Questions to the Prime Minister, have gained renown for their partisan clashes, often fueled by the (mis)handling of a current crisis. The following question, asked to the Prime Minister by an opposition MP about contamination of the meat supply in 2013, encapsulates this odd mix of purposes:

{ \em ``The Prime Minister is rightly shocked by the revelations that many food products contain 100\% horse. Does he share my concern that, if tested, many of his answers may contain 100\% bull?"}\footnote{MPs almost always  
 address each other in 3rd person.}

The moderated, relatively rigid format of questions periods, along with the multifaceted array of underlying incentives and interpersonal relationships, yields a structurally controlled setting with a rich variety of social interactions, taking place in the realm of important policy discussions. This complexity makes question periods a particularly fruitful and consequential setting in which to study questions as social signals, and expand our understanding of their role beyond factual queries.

 \xhdr{Dataset description} 
Our dataset covers question periods from May 1979 to December 2016, encompassing six different Prime Ministers.
For each question period, we extract all question-answer pairs, along with the identity of the asker and answerer.
Because our focus here is on how questions are posed in a social setting, and not on the subsequent dialogue, we ignore questions which were tabled prior to the session, as well as any followup back-and-forth dialogue between the asker and answerer.  

We augment this collection with metadata about each asker and answerer, including their political party, the time when they first took office, and whether they were serving as a minister 
at a given point in time.  Such information is used to validate our methodology and interpret our results in light of the social context in which the questions were asked, described further in Sections \ref{sec:validation} and \ref{sec:insights}.

In total there are 216,894 question-answer pairs in our data, occurring over 4,776 days and 6 prime-ministerships. The questions cover 1,975 different askers, 1,066 different answerers, and a variety of government departments with responsibilities ranging from defense to transport.
We make this dataset publicly available, along with the code implementing our methodology, as part of the Cornell Conversational Analysis Toolkit.\footnote{\scriptsize{\url{https://github.com/CornellNLP/Cornell-Conversational-Analysis-Toolkit}}}

\section{Question Motifs}
\label{sec:motifs}
The first component of our framework identifies lexico-syntactic phrasing patterns recurring in a collection of questions, which we call {\em motifs}.
Intuitively, motifs constitute wordings commonly used to pose questions. 
To find motifs in a given collection, we first extract relevant {\em fragments}
 from each question. We then 
 group
  sets of frequently co-occurring fragments into motifs.

\xhdr{Question fragments}
Our goal is to find motifs which reflect functional characteristics of questions. Hence, we start by extracting the key {\em fragments} within a question which encapsulate its functional nature. Following the intuition that the bulk of this functional information is contained in the root of a question's dependency parse along with its outgoing arcs \cite{iyyer2014neural}, we take the fragments of a question to be the root of its parse tree, along with each (root, child) pair. To capture cases when the operational word in the question is not connected to its root (such as ``What...''), we also consider the initial unigram and bigram of a question as fragments.  The following question has 5 fragments: what, what is, going$\rightarrow$*, is$\leftarrow$going and going$\rightarrow$do.

{\footnotesize
\begin{dependency}[theme = simple]
   \begin{deptext}[column sep=0.1em]
    \hspace{-0.6cm} (1) \ \ \ \ \ \textbf{What} \& \textbf{is} \& the minister \& \textbf{going} \& to \& \textbf{do} \& about \&  ... ? \\
   \end{deptext}
   \depedge[arc angle=30]{4}{2}{}
   \depedge[arc angle=30]{4}{6}{}
\end{dependency}
}

Because 
our goal is to capture topic-agnostic patterns, we ignore all fragments which contain a noun phrase (NP) or pronoun. 
NP subtrees are identified based on their outgoing dependencies to the root;\footnote{We take as NPs subtrees connected to the root with the following: nsubj, nsubjpass, dobj, iobj, pobj, attr.} 
in the event that an NP starts with a WH-determiner (WDT), we consider (root, WDT) to be a fragment and drop the remainder of the NP.\footnote{In the particular case of the Parliament dataset, removing NPs also removes conventional, partisan address terms (e.g. ``my hon. Friend'').} 

Finally, we note that some questions consist of multiple sub-questions (``What does the Minister think [...], {\em and} why [...]?''). For such questions, 
we recursively extract fragments from each child subtree in the same manner, starting from their roots.

\xhdr{From fragments to motifs}
We define \textit{motifs} as sets of question fragments that frequently co-occur (in at least $n$ questions). We find motifs by applying the apriori algorithm \cite{agrawal1994fast} to find these common itemsets.  
This results in a collection of motifs $\mathcal{M}$ which correspond to different question phrasings.\footnote{In some cases, a pair of motifs almost always co-occurs in the same questions, making them redundant. We treat two motifs $m_1$ and $m_2$ as equivalent if, for some probability $p$, $\Pr(m_1 | m_2) > p$ and $\Pr(m_2 | m_1) > p$; we keep the smaller of the two as the representative motif, or pick one of them arbitrarily if they are of equal sizes.}
Examples of motifs are shown in Table \ref{tab:qtypes}.

Motifs can identify phrasings to varying degrees of specificity. For example, the singleton motif \{what is\} corresponds to all questions starting with that bigram, while \{what is, going$\rightarrow$do\} narrows these down to questions also containing the fragment going$\rightarrow$do.   To model the specificity relation between motifs, we structure $\mathcal{M}$ as a directed acyclic graph where an edge points from a motif $m_1$ to another motif $m_2$ if the latter has exactly one more fragment in addition to those in $m_1$, corresponding to a narrower set of phrasings.

\xhdr{Motif-representation of a question}
Finally, a question $q$ contains a motif if it includes all of the fragments comprising that motif. We can hence capture the phrasing of a given question $q$ using the subset of motifs it contains, structured as the subgraph $\mathcal{M}_q \subset \mathcal{M}$ induced by this subset.  This directed subgraph represents the question at multiple levels of specificity simultaneously; in particular, the set of sinks (i.e., nodes with outdegree 0; henceforth {\em sink motifs}) of $\mathcal{M}_q$ is the most fine-grained way to specify the phrasing of $q$.   For example \{what is, is$\leftarrow$going, going$\rightarrow$do\} is the only sink motif of the question in example (1); its entire subgraph is shown in Figure \ref{fig:subgraph} in the appendix. 

\section{Latent Question Types}
\label{sec:qtypes}
The second component of our framework structures the space of questions according to their functional roles, thus going beyond the lexico-syntactic representation captured via motifs.    The main intuition is that the nature of the answer that a question receives provides a good indication of its intention.  
Therefore, if two questions are phrased differently but answered in similar ways, the parallels exhibited by their answers should reflect commonalities in the askers' intentions.
To operationalize this intuition, we first construct a latent space based on answers, and  
then map question motifs (Section \ref{sec:motifs}) to the same space. Using the resultant latent representations, we can then %
cluster questions with similar rhetorical functions, even if their surface forms are different.

\xhdr{Constructing a space of answers}
In line with our focus on functional characterizations, we extract the fragments from each sentence of an \textit{answer}, defined in the same way as question fragments. 
We then construct a term-document matrix, where terms correspond to answer fragments, and documents correspond to individual answers in the corpus. We filter out infrequent fragments occurring less than $n_A$ times,  reweight the rows of this matrix with tf-idf reweighting, and scale to unit norm, producing a fragment-answer matrix $\mathcal{A}$. We perform singular value decomposition on $\mathcal{A}$ and obtain a low-rank representation  $\mathcal{A} \approx \hat{\mathcal{A}} = U_A S V_A^T$, for some rank $d$, where rows of $U_A$ correspond to answer fragments and rows of $V_A$ correspond to answers.\footnote{We experimented with grouping answer fragments into motifs as well, but found that most of the motifs produced were one fragment large. While future work could focus more on understanding consistent phrasings of answers, we note that at least in our chosen corpus, answers are longer and encompass a much greater variation of possible phrasings.%
} 

\xhdr{Latent projection of question motifs}
We can draw a natural correspondence between a question motif $m$ and answer term $t$ if $m$ occurs in a question whose answer contains $t$. This enables us to compute representations of question motifs in the same space as $\hat{\mathcal{A}}$. 
Concretely, we construct a motif-question matrix $\mathcal{Q} = (q_{ij})$ where $q_{ij} = 1$ if motif $i$ occurred in question $j$; we scale rows of $\mathcal{Q}$ %
to unit norm. To represent $\mathcal{Q}$ in the latent answer space, we solve for $\hat{\mathcal{Q}}$ in $\mathcal{Q} = \hat{\mathcal{Q}}SV_A^T$ as $\hat{\mathcal{Q}} = \mathcal{Q}V_A S^{-1}$, again scaling rows to unit norm. 
 Row $i$ of $\hat{\mathcal{Q}}$ then gives a $d$-dimensional representation of motif $i$, denoted $\hat{q_i}$. %

\xhdr{Grouping similar questions}
Finally, we identify {\em question types}---broad groups of similar motifs. Intuitively, if two motifs $m_i$ and $m_j$ have vectors $q_i$ and $q_j$ which are close together, 
they elicit answers that are close in the latent space, so are functionally similar in this sense.
We use the K-Means algorithm \cite{scikit-learn} to cluster motif vectors into $k$ clusters; these clusters then constitute the desired set of question types. 

To determine the type of a {\em particular} question $q^*$, we transform it to a binary vector ($q^*_i$) where $q^*_i = 1$ if motif $i$ is a sink motif of $q^*$; using only sink motifs at this stage allows us to characterize a question according to the most specific representation of its phrasing, thus avoiding spurious associations resulting from more general motifs.
We scale $q^*$, project it to the latent space as before, and assign the resultant projection $\hat{q}^*$ to a cluster $t$, hence determining its type.

Since question motifs and answer fragments have both been mapped to the same latent space (as rows of $\hat{\mathcal{Q}}$ and $U_A$ respectively), we can also assign each answer fragment to a question type. 
This further facilitates interpretability through characterizing the answers commonly triggered by a particular type of question.

\begin{table*}[t]
\begin{center}
\begin{tabular}{p{1.1cm}  p{2.6cm}  p{2.4cm}  p{8.35cm} }
\multicolumn{1}{c}{\footnotesize \textbf{Type}} & \multicolumn{1}{c}{\footnotesize \textbf{Question motifs} }& \multicolumn{1}{c}{\footnotesize \textbf{Answer fragments} }& \multicolumn{1}{c}{\small \textbf{Example question-answer pairs} }\\
\hline 
\footnotesize \multirow{2}{1.5cm}{\footnotesize 0: Issue update (16,693)}	& \multirow{2}{4cm}{\footnotesize \{what,\hspace{0.1em}are$\leftarrow$taking\}, \{will$\leftarrow$update\}} 	& \multirow{2}{2cm}{\footnotesize continue$\rightarrow$work, met$\rightarrow$discuss} & \footnotesize Q: \textbf{What} steps \textbf{are} the Department \textbf{taking} to create a system for asylum-seekers?\\[-2pt]
									&&& \footnotesize A: We \textbf{continue} to \textbf{work} with the Department of Education to ensure an equitable  [...]\\[2pt]
\hline 
\footnotesize \multirow{2}{1.5cm}{\footnotesize 1: Shared concerns (35,954)}	& \multirow{2}{2cm}{\footnotesize \{will$\leftarrow$take\}, \{may$\leftarrow$urge\}} 	& \multirow{2}{2cm}{\footnotesize grateful$\leftarrow$am, shall$\leftarrow$consider} & \footnotesize Q: \textbf{Will} he \textbf{take} steps to support other MPs to employ apprentices?\\[-2pt]
									&&& \footnotesize A: I \textbf{am grateful} for that suggestion [...]\\[8pt]

\hline 
\footnotesize \multirow{2}{1.5cm}{\footnotesize 2: Narrow factual (16,467)}	& \multirow{2}{2cm}{\footnotesize \{what$\leftarrow$made\}, \{what$\leftarrow$happen, will$\leftarrow$happen\}} 	& \multirow{2}{2cm}{\footnotesize is$\leftarrow$considering, have$\leftarrow$discussed} & \footnotesize Q: \textbf{What} representations has the Minister \textbf{made} on the future of rural policing [in] Dyfed-Powys?\\[-4pt]
									&&& \footnotesize A:The Home Office \textbf{is considering} the matter [...]\\[2pt]	
\hline 
\footnotesize \multirow{2}{1.5cm}{\footnotesize 3: Prompt for comment (16,588)}	& \multirow{2}{4cm}{\footnotesize \{what$\leftarrow$say,\hspace{0.1em}say$\rightarrow$to\}, \{will$\leftarrow$tell\}}  	& \multirow{2}{2cm}{\footnotesize must$\leftarrow$say, said$\rightarrow$was } & \footnotesize Q: \textbf{What} has the Prime Minister to \textbf{say to} President Reagan for sending troops to Honduras? \\[-2pt]
									&&& \footnotesize A: [...] I \textbf{must say} that we deplore the reported incursion by Nicaraguan forces [...] \\
									
\hline 
\footnotesize \multirow{2}{1.5cm}{\footnotesize 4:~Agreement (32,835)}	& \multirow{2}{2cm}{\footnotesize \{does$\leftarrow$agree, agree$\rightarrow$is\}, \{is$\rightarrow$important\}} 	& \multirow{2}{2cm}{\footnotesize agree$\rightarrow$with, agree$\rightarrow$completely} & \footnotesize Q: \textbf{Does} [he] \textbf{agree} that one of the best ways to improve the trade balance \textbf{is} to continue the Government's strong economic policies? \\[-2pt]
									&&& \footnotesize A: I \textbf{agree with} my hon. Friend [...]\\
\hline 
\footnotesize \multirow{2}{1.5cm}{\footnotesize 5: Self promotion (26,351)}	& \multirow{2}{2cm}{\footnotesize \{is$\rightarrow$aware\}, \{will$\leftarrow$consider\}} 	& \multirow{2}{2cm}{\footnotesize will$\leftarrow$appreciate, am$\rightarrow$certain} & \footnotesize Q: \textbf{Is} my Friend \textbf{aware} that members of my parish church are pleased to have received a grant [...] ? \\[-2pt]
									&&& \footnotesize A: [My Friend] \textbf{will appreciate} the significant performance of parishes up and down the country [...] \\
\hline 
\footnotesize \multirow{2}{1.5cm}{\footnotesize 6:~Concede, accept (31,653)}	& \multirow{2}{2cm}{\footnotesize \{will$\leftarrow$accept\}, \{is$\rightarrow$not,~is$\rightarrow$true\}} 	& \multirow{2}{2cm}{\footnotesize not$\leftarrow$accept,  not$\leftarrow$believe} & \footnotesize Q: \textbf{Will} [he] \textbf{accept} that [the UK exiting the EU] would undermine our security  [...]?\\[-2pt]
									&&& \footnotesize A: No, I do \textbf{not accept} that [...]\\
\hline 
\footnotesize \multirow{2}{1.5cm}{\footnotesize 7: Condemnatory (21,320)}	& \multirow{2}{2cm}{\footnotesize \{can$\leftarrow$explain\}, \{how$\leftarrow$justify, can$\leftarrow$justify\}} 	& \multirow{2}{2cm}{\footnotesize knows$\rightarrow$well, is$\rightarrow$wrong} & \footnotesize Q: \textbf{Can} the Secretary \textbf{explain} why the Government are scrapping child poverty targets?\\[-2pt]
									&&& \footnotesize A: The hon. Lady \textbf{is wrong} in what she says [...]\\	
\end{tabular}
\end{center}
\vspace{-0.1cm}
 \caption{Question types automatically extracted from the parliamentary question periods, %
 along with representative motifs and question-answer pairs. 
The number of questions in our dataset assigned to each type is shown in parantheses.
  Interpretations and more examples in Tables \ref{tab:appendix1} \& \ref{tab:appendix2} in the appendix. 
 }
 \label{tab:qtypes}
\end{table*}

\section{Validation}
\label{sec:validation}
We now apply our general framework to the particular setting of parliamentary question periods, structuring the space of questions posed within these sessions according to their rhetorical function. To validate the induced typology, we quantitatively show that it recovers asker intentions in an expert-coded dataset, and qualitatively aligns with prior findings in the political science literature on parliamentary dynamics.
\begin{figure*}[t]
\includegraphics[width=1\textwidth]{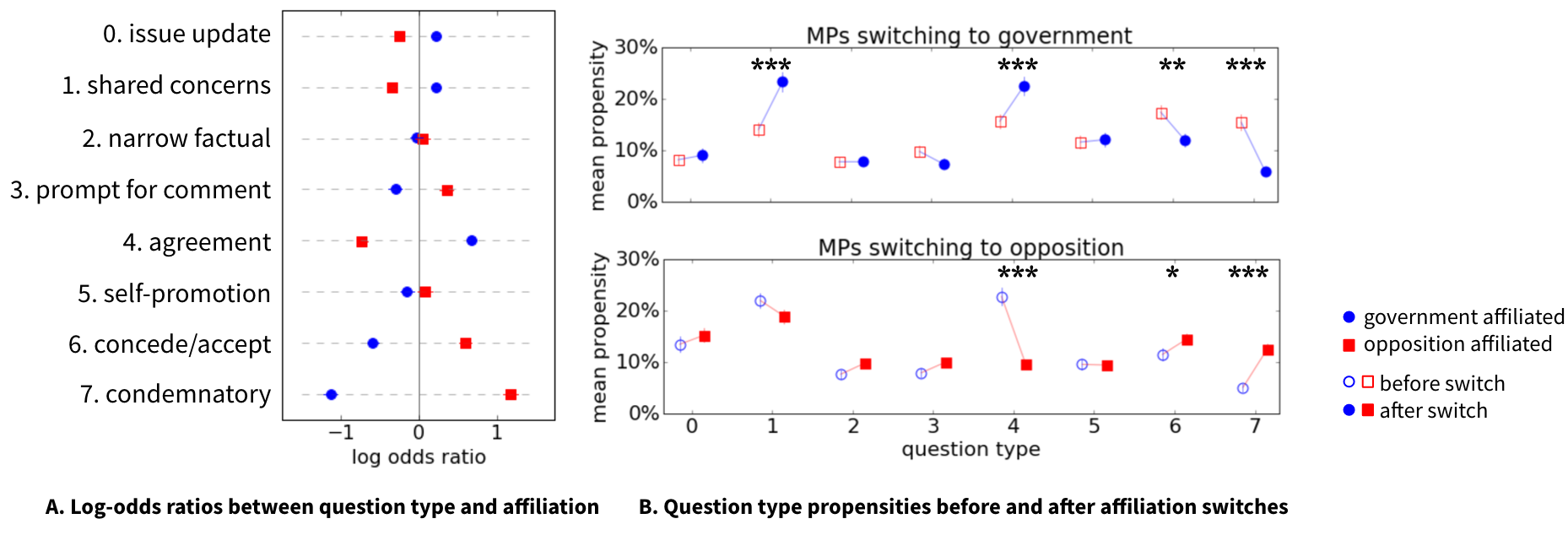}
\caption{
\textbf{A}: Log-odds ratios of questions of each type asked by government and opposition MPs, compared to MPs not of the respective affiliation; 95\% confidence intervals (sometimes imperceptible) are depicted. 
\textbf{B}: Mean propensities for each question type, for MPs who switch from being in the opposition to being in the government (top) and vice-versa (bottom) after an election. Stars indicate statistically significant differences at the $p < 0.05$ (*), $p < 0.01$ (**) and $p < 0.001$ (***) levels (Wilcoxon test).
}
\label{fig:qtype_party}
\end{figure*}

\xhdr{Question types in Parliament} We apply our motif extraction and question type induction pipeline to the questions in the parliamentary dataset.\footnote{We consider questions to be sentences ending in question marks. If an utterance consists of multiple questions, we extract fragments sets from each question separately, and take the motifs of the utterance to be the union of motifs of each component question. We set $n = 100$, $p = 0.9$, $n_A=100$ and $d=25$. The choice of parameters was done via manual inspection of the dataset.}
Over 90\% of the questions in the dataset contain at least one of the resulting 2,817 motifs; in subsequent analyses we discard questions without a matching motif.
We apply our pipeline to the questions in the parliamentary dataset, and induce a typology of $k=8$ question types to capture the rich array of questions represented in this space while preserving interpretability. 

Table \ref{tab:qtypes} 
displays extracted types, along with example questions, answers, and motifs.\footnote{Each type contains a few hundred question motifs and answer fragments.}
   The second author, a political scientist with domain expertise in the UK parliamentary setting, manually investigated each type and provided interpretable labels.  
For example, in questions of type 4, the asker is aware that his main premise is supported by the minister, and thus will be met with a positive statement backing the thrust of the question; we call this the \textbf{agreement} cluster. Types 6 and 7 are much more combative: in type 6  questions the asker explicitly attempts to force the minister to \textbf{concede/accept} a point that would undermine some government stance, while type 7 contains \textbf{condemnatory} questions that prompt the minister to justify a policy that is self-evidently bad in the eyes of the asker. In contrast, type 2 constitutes tamer \textbf{narrow} queries that require the minister to simply report on non-partisan matters of policy. (Extended interpretations in the appendix.)
\xhdr{Quantitative validation}
We compare our output to a dataset of 1,256 questions asked to various Prime Ministers labeled by \citet{Batesetal2014} (also included in our data distribution). 
Each question in this data is hand-coded by a domain expert with one of three labels %
indicating the rhetorical intention of the asker: compared to {\em standard} questions---denoting straightforward factual queries, {\em helpful} questions serve as prompts for the PM to talk favorably about their government, while {\em unanswerable} questions are effectively vehicles for delivering criticisms that the PM cannot respond to. Questions which are {\em unanswered} by the PM are also labeled.
If our framework captures meaningful rhetorical dimensions, we expect a given label to be over-represented in some of our induced types, and under-represented in others.

Even though our clustering of questions is generated in an unsupervised fashion without any  guidance from the coded rhetorical roles, we see that several of the types we discover closely align with these annotations. In particular, helpful questions are highly associated with the \textbf{agreement} type (constituting 28\% of questions of that type compared to 14\% over the entire dataset; binomial test $p < 0.01$),
reinforcing our interpretation that this type captures MPs cheerleading their own government. Conversely, unanswerable questions are frequently of the \textbf{concede/accept} type (20\% in-type vs.~11\% overall), while \textbf{condemnatory} questions are often unanswered (43\% vs.~24\% overall),  suggesting that questions of these types have an increased tendency to be posed as aggressive criticisms packaged as questions.%
We also validate our framework in a prediction setting using these labels, in three binary classification tasks: distinguishing %
helpful  vs.~standard, unanswerable vs.~standard, and unanswered vs.~answered questions. (In each task, we balance the two classes.)
To control for asker affiliation effects, we consider only questions asked by government MPs for the helpful task, and opposition questions in the unanswerable and unanswered tasks; we train on questions to Conservative PMs and evaluate on Labour PMs.\footnote{These choices are motivated by the number of questions from each affiliation and party in the dataset (see appendix for further details on this dataset).}
For each setting, we train logistic regression classifiers; as features we compare the latent representation of each question to a unigram BOW baseline.\footnote{We used tf-idf reweighting and excluded unigrams occurring less than 5 times.}

In the {\em unanswerable} and {\em unanswered} tasks, we find that the BOW features do not perform significantly better than a random (50\%) baseline. However, the latent question features produced by our framework bring additional predictive signal and outperform the baseline when combined with BOW (binomial $p < 0.05$), achieving accuracies of 66\% and 62\% respectively (compared with 55\% and 50\% for BOW alone). 
This suggests that our representation captures useful rhetorical information that, given our train-test split, generalizes across parties. None of the models significantly outperform the random baseline on the {\em helpful} task, perhaps owing to the small data size. 

\begin{figure*}
\includegraphics[width=1\textwidth]{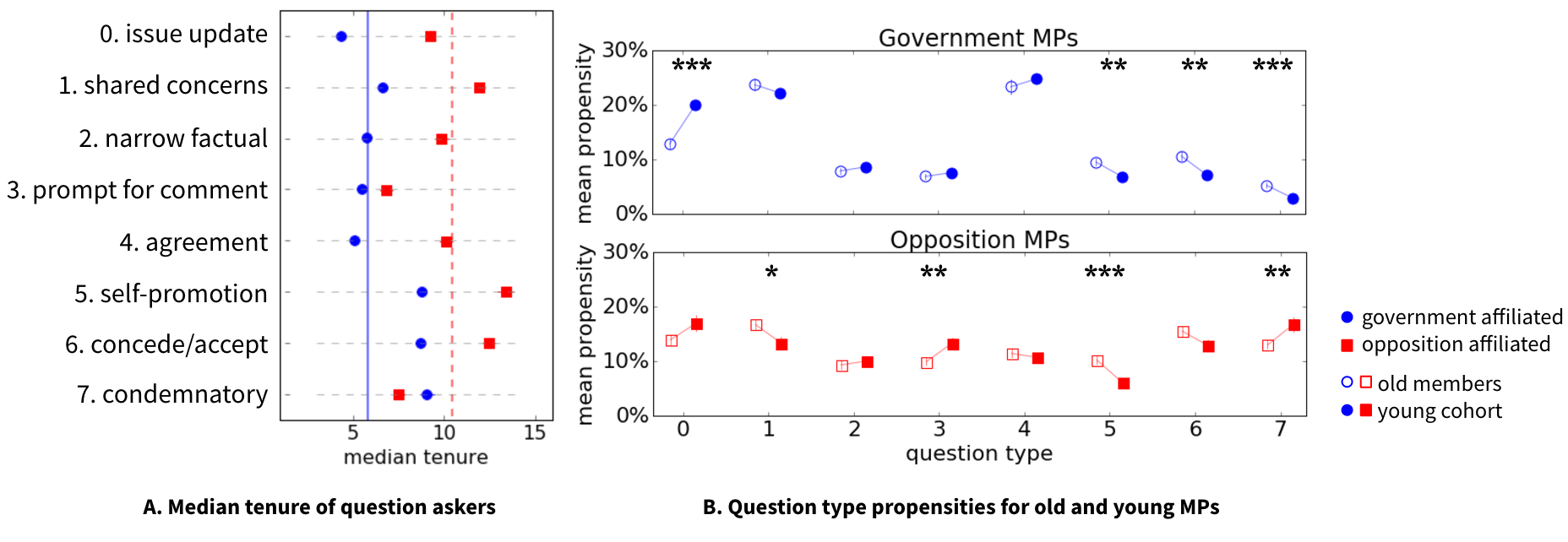}
\caption{
\textbf{A}: Median asker tenures over each question type, for government and opposition askers. Overall median tenures are also shown for reference (solid blue line for government, dashed red line for opposition).
\textbf{B}: Mean propensities for newly elected MPs during the 1997 and 2010 elections, compared to re-elected MPs in the subsequent parliamentary sitting. 
Stars indicate statistically significant differences at the $p < 0.05$ (*), $p < 0.01$ (**) and $p < 0.001$ (***) levels (Mann Whitney U test).}
\label{fig:tenure}
\end{figure*}
\xhdr{Qualitative validation: question partisanship}
We additionally provide a qualitative validation of our framework by comparing the question-asking activity of government and opposition-affiliated MPs---as viewed through the extracted question types---to well-established characterizations of these affiliations in the political science literature. In particular, prior work has examined the bifurcation in behavior between government and opposition members, in their differing focus on various issues \cite{louwerse2012mechanisms}, and in settings such as roll call votes \cite{Cowley02,SpirlingMcLean07,EggersSpirling2012}. Since government MPs are elected on the same party ticket and manifesto, they primarily act to support the government's various policies and bolster the status of their cabinet, seldom airing disagreements publicly. In contrast, opposition members tend to offer trenchant partisan criticism of government policies, seeking to destabilize the government's relationship with its MPs and create negative press in the country at large. 
In characterizing the question-asking activity of government and opposition MPs, this friendly vs.~adversarial behavior should also be reflected in a rhetorical typology of questions.\footnote{While we induce the typology over our entire dataset, we perform all subsequent analyses on a filtered subset of 50,152 questions. In particular, we omit utterances with multiple questions---i.e. multiple question marks---to ensure that we don't confound effects arising from different co-occurring question types. Our filtering decisions are also determined by the availability of information about the asker and answerers' roles in Parliament. Further information about these filtering choices can be found in the appendix.}

Concretely, to quantify the relationship between a particular question type $t$ and asker affiliation $\mathcal{P}$, we compute the log-odds ratio of type $t$ questions asked by MPs in $\mathcal{P}$, compared to MPs not in $\mathcal{P}$.\footnote{The log-odds values are not symmetric between government and opposition, because they includes questions asked by MPs not in the official opposition.}

Figure \ref{fig:qtype_party}A shows the resultant log-odds ratios of each question type for government and opposition members. Notably, we see that \textbf{agreement}-type questions are significantly more likely to originate from government than from opposition MPs, while the opposite holds for \textbf{concede/accept} and \textbf{condemnatory} questions (binomial $p < 10^{-4}$ for each, comparing within-type to overall proportions of questions from an affiliation). No such slant is exhibited in the \textbf{narrow factual} type, further reinforcing the role of such questions as informational queries about relatively non-partisan issues. These results strongly cohere with the ``textbook'' accounts of parliamentary activity in the literature, as well as our interpretation of these types as bolstering or antagonistic.%

Moreover, we find that the {\em same MP} shifts in her propensity for different question types as her affiliation changes. When a new political party is elected into office, MPs who were previously in the opposition now belong to the government party, and vice versa. 
Such a switch occurs within our data between the Major and Blair governments (Conservative to Labour, 1997), and between the Brown and Cameron governments (Labour to Conservative, 2010). For both switches, we consider all MPs who asked at least 5 questions both before and after the switch, resulting in 88 members who {\em became} government MPs and 102 who became opposition MPs.
For an MP $M$ we compute $P_{M,t}$, their {\em propensity} for a question type $t$, as the proportion of questions they ask which are from $t$. Comparing $P_{M,t}$ before and after a switch, we replicate the key differences observed above---for instance, we find that former opposition MPs who become government MPs decrease in their propensity for \textbf{condemnatory} questions, while newly opposition MPs move in the other direction (Wilcoxon $p < 0.001$, Figure \ref{fig:qtype_party}B). This  suggests that the general trends we observed before are driven by the shift in affiliation, and hence parliamentary role, of {\em individual} MPs.%

\section{Career Trajectory Effects} %
\label{sec:insights}
We now apply our framework to gain further insights into the nature of political discourse in Parliament, focusing on how questioning behavior varies with a member's tenure in the institution. As stated in the introduction, two alternative hypotheses arise: younger
MPs %
may be more vigorously critical out of enthusiasm, but are potentially tempered by their stake in future promotion prospects compared to older members \cite{Cowley02,cowley2012arise}.
Alternatively, older MPs who have less at stake in terms of prospects of further promotion may ask more antagonistic questions. 
Throughout, \textit{young} and \textit{old} refer to tenure---i.e., how many years someone has served as an MP---rather than biological age.

In order to understand the extent to which young or old members contribute a specific type of question, for each question type $t$ we compute the median tenure of askers of each question in $t$, and compare the median tenures of different question types, for each affiliation (Figure \ref{fig:tenure}A).\footnote{Median
 tenures for opposition members are generally higher; winning an election tends to result in more newly-elected and therefore younger MPs \cite{webb1999party}.}
 We see that among both affiliations, more aggressive questions tend to originate more from older members, reflected in significantly higher median tenures (for types 6 in both affiliations, and 7 in government MPs; Mann Whitney U test $p < 0.001$ comparing within-type median tenure with outside-type median tenure); whereas standard \textbf{issue update} questions tend to come from younger members ($p < 0.001$, both affiliations). Notably, the disproportionate aggressiveness of older members manifests even among {\em government} MPs who direct these questions towards their {\em own} government. This supports the ``less to lose'' intuition, offering a rhetorical parallel to 
 previous findings about the increased tendency to vote contrary to party lines from MPs with little chance of ministerial promotion \cite{BenedettoHix07}.

Interestingly, we find that these differential preferences across member tenure also manifest at a finer granularity than simply less to more aggressive. %
For instance, younger {\em opposition} members tend to contribute more \textbf{condemnatory} questions compared to older members (Mann Whitney U test $p < 0.01$), who disproportionately favor \textbf{concede/accept} questions. While further work is needed to fully explain these differences, we speculate that they are potentially reflective of strategic attempts by younger MPs to signal traits that could facilitate future promotion, such as partisan loyalty \cite{Kam09}.

To discount the possibility of these effects being solely driven by a few very prolific young or old MPs, we also consider a setting where type propensities are macroaveraged over MPs. For each affiliation we compare the cohort of younger MPs who are newly voted in at the 1997 and 2010 elections, with older MPs who have been in office prior to the election.\footnote{This totals 272 new and 184 old government MPs, and 84 new and 179 old opposition MPs.} We compute the type propensities of these two cohorts over the questions they asked during the \textit{subsequent} parliamentary sitting,
and replicate the tenure effects observed previously (Figure \ref{fig:tenure}B). This suggests that these parliamentary career effects reflect behavioral changes at the level of individual MPs,  whose incentives evolve over their tenure.

\section{Conclusion and Future Work}
\label{sec:discussion}
In this work we introduced an unsupervised framework for structuring the space of questions according to their rhetorical role.  
We instantiated and validated our approach in the domain of parliamentary question periods, and revealed new interactions between questioning behavior and career trajectories.

We note that our methodology is not tied to a particular domain.  It would be interesting to explore its potential in a variety of less structured domains where questions likewise play a crucial role.  For example, examining how interviewers in high-profile media settings (e.g., Frost on Nixon) can use their questions to elicit substantive responses from influential people would aid us in the broader normative goal of holding elites to account, by gaining a better understanding of what and how to ask, and what (not) to accept as an answer.%

From a technical standpoint, future work could also augment the representation of questions and answers presently used in our framework, beyond our heuristic of using root arcs without noun phrases. Richer linguistic representations, as well as more judicious ways of weighting different fragments and motifs, could enable us to capture a wider range of possible surface and rhetorical forms, especially in settings where phrasings are potentially less structured by institutional conventions.  Additionally, as with most unsupervised methods, our approach is limited by the need to hand-select parameters such as the number of clusters, and manually interpret the typology's output. Having annotations of these corpora could better motivate the methodology and enable further evaluation and interpretation; we hope to encourage such annotation efforts by releasing the dataset.

Inevitably, drawing causal lessons from observational data is difficult. Moving forward,
experimental tests of insights gathered through such explorations would enable us to establish causal effects of question-asking rhetoric, perhaps offering prescriptive insights into questioning strategies for objectives such as information-seeking \cite{dillman1978mail}, request-making \cite{Althoff+al:14a,Mitra:ProceedingsOfCscw:2014} and persuasion \cite{tan2016winning,Zhang:Naacl:2016,wangwinning}. 
{\small
\xhdr{Acknowledgements}
The first author thanks John Bercow, the Speaker of the House, for suggesting she ``calm [her]self down by taking up yoga'' during the hectic deadline push (\url{https://youtu.be/AiAWdLAIj3c}).  
The authors thank the anonymous reviewers and Liye Fu for their comments and for their \textit{helpful} questions.  We are grateful to the organizers of the conference on New Directions in Text as Data for fostering the inter-disciplinary collaboration that led to this work, to Amber Boydstun and Philip Resnik for  their insights on questions in the political domain, and to Stephen Bates, Peter Kerr and Christopher Byrne for sharing the labeled PMQ dataset.  This research has been supported in part by a Discovery and Innovation Research Seed Award from the Office of the Vice Provost for Research at Cornell.
}

\bibliography{paper-questions}

\begin{thebibliography}{51}
\expandafter\ifx\csname natexlab\endcsname\relax\def\natexlab#1{#1}\fi

\bibitem[{Agrawal and Srikant(1994)}]{agrawal1994fast}
Rakesh Agrawal and Ramakrishnan Srikant. 1994.
\newblock Fast algorithms for mining association rules.
\newblock In \emph{Proceedings of VLDB}.

\bibitem[{Althoff et~al.(2014)Althoff, Danescu-Niculescu-Mizil, and
  Jurafsky}]{Althoff+al:14a}
Tim Althoff, Cristian Danescu-Niculescu-Mizil, and Dan Jurafsky. 2014.
\newblock How to ask for a favor: A case study on the success of altruistic
  requests.
\newblock In \emph{Proceedings of ICWSM}.

\bibitem[{Aubakirova and Bansal(2016)}]{Aubakirova:ProceedingsOfEmnlp:2016}
Malika Aubakirova and Mohit Bansal. 2016.
\newblock Interpreting neural networks to improve politeness comprehension.
\newblock In \emph{Proceedings of EMNLP}.

\bibitem[{Bates et~al.(2014)Bates, Kerr, Byrne, and Stanley}]{Batesetal2014}
Stephen~R. Bates, Peter Kerr, Christopher Byrne, and Liam Stanley. 2014.
\newblock Questions to the {Prime} {Minister}: A comparative study of {PMQs}
  from {Thatcher} to {Cameron}.
\newblock \emph{Parliamentary Affairs}.

\bibitem[{Benedetto and Hix(2007)}]{BenedettoHix07}
Giacomo Benedetto and Simon Hix. 2007.
\newblock The {Rejected}, the {Ejected}, and the {Dejected}: {Explaining}
  government rebels in the 2001-2005 {British House of Commons}.
\newblock \emph{Comparative Political Studies}.

\bibitem[{Bhattasali et~al.(2015)Bhattasali, Cytryn, Feldman, and
  Park}]{bhattasali2015automatic}
Shohini Bhattasali, Jeremy Cytryn, Elana Feldman, and Joonsuk Park. 2015.
\newblock Automatic identification of rhetorical questions.
\newblock In \emph{ACL}.

\bibitem[{Boydstun et~al.(2014)Boydstun, Card, Gross, Resnik, and
  Smith}]{Boydstun:ProceedingsOfTheApsa:2014}
Amber~E. Boydstun, Dallas Card, Justin~H. Gross, Philip Resnik, and Noah~A.
  Smith. 2014.
\newblock {Tracking the development of media frames within and across policy
  issues}.
\newblock In \emph{Proceedings of the APSA}.

\bibitem[{Bull and Wells(2012)}]{BullWells2012}
Peter Bull and Pam Wells. 2012.
\newblock Adversarial discourse in {Prime Minister’s Questions}.
\newblock \emph{Journal of Language and Social Psychology}.

\bibitem[{Cao et~al.(2010)Cao, Cong, Cui, and Jensen}]{cao2010generalized}
Xin Cao, Gao Cong, Bin Cui, and Christian~S Jensen. 2010.
\newblock A generalized framework of exploring category information for
  question retrieval in community question answer archives.
\newblock In \emph{Proceedings of WWW}.

\bibitem[{Card et~al.(2016)Card, Gross, Boydstun, and Smith}]{Card:Emnlp:2016}
Dallas Card, Justin Gross, Amber Boydstun, and Noah Smith. 2016.
\newblock Analyzing framing through the casts of characters in the news.
\newblock In \emph{Proceedings of EMNLP}.

\bibitem[{Chester and Bowring(1962)}]{ChesterBowring1962}
Daniel Chester and Nona Bowring. 1962.
\newblock \emph{Questions in Parliament}.
\newblock Clarendon Press, Oxford.

\bibitem[{Cowley(2002)}]{Cowley02}
Philip Cowley. 2002.
\newblock \emph{The Rebels: How Blair Mislaid his Majority}.
\newblock Politico's Publishing.

\bibitem[{Cowley(2012)}]{cowley2012arise}
Philip Cowley. 2012.
\newblock Arise, novice leader! {The} continuing rise of the career politician
  in {Britain}.
\newblock \emph{Politics}.

\bibitem[{Danescu-Niculescu-Mizil et~al.(2013)Danescu-Niculescu-Mizil, Sudhof,
  Jurafsky, Leskovec, and Potts}]{Danescu-Niculescu-Mizil+al:13b}
Cristian Danescu-Niculescu-Mizil, Moritz Sudhof, Dan Jurafsky, Jure Leskovec,
  and Christopher Potts. 2013.
\newblock A computational approach to politeness with application to social
  factors.
\newblock In \emph{Proceedings of ACL}.

\bibitem[{Dhillon et~al.(2004)Dhillon, Sonali, Hannah, and
  Elizabeth}]{dhillon2004meeting}
Rajdip Dhillon, Bhagat Sonali, Carvey Hannah, and Shriberg Elizabeth. 2004.
\newblock {Meeting Recorder Project: Dialog Act Labeling Guide}.
\newblock Technical report.

\bibitem[{Dillman(1978)}]{dillman1978mail}
Don~A Dillman. 1978.
\newblock \emph{Mail and telephone surveys: The total design method}.
\newblock Wiley New York.

\bibitem[{Eggers and Spirling(2014)}]{EggersSpirling2012}
Andrew Eggers and Arthur Spirling. 2014.
\newblock Ministerial responsiveness in westminster systems: Institutional
  choices and {House of Commons} debate, 1832--1915.
\newblock \emph{American Journal of Political Science}.

\bibitem[{Goffman(1976)}]{goffman1976replies}
Erving Goffman. 1976.
\newblock {Replies and responses}.
\newblock \emph{Language in society}.

\bibitem[{Gonzalez-Bailon et~al.(2010)Gonzalez-Bailon, Kaltenbrunner, and
  Banchs}]{gonzalez2010structure}
Sandra Gonzalez-Bailon, Andreas Kaltenbrunner, and Rafael~E Banchs. 2010.
\newblock {The structure of political discussion networks: A model for the
  analysis of online deliberation}.
\newblock \emph{Journal of Information Technology}.

\bibitem[{Grimmer et~al.(2012)Grimmer, Messing, and
  Westwood}]{Grimmer:AmericanPoliticalScienceReview:2012}
Justin Grimmer, Solomon Messing, and Sean~J Westwood. 2012.
\newblock How words and money cultivate a personal vote: {The} effect of
  legislator credit claiming on constituent credit allocation.
\newblock \emph{American Political Science Review}.

\bibitem[{Grimmer and Stewart(2013)}]{Grimmer:PoliticalAnalysis:2013}
Justin Grimmer and Brandon Stewart. 2013.
\newblock {Text as Data: The} promise and pitfalls of automatic content
  analysis methods for political texts.
\newblock \emph{Political Analysis}.

\bibitem[{Gupta and Gupta(2012)}]{gupta2012survey}
Poonam Gupta and Vishal Gupta. 2012.
\newblock A survey of text question answering techniques.
\newblock \emph{International Journal of Computer Applications}.

\bibitem[{Harabagiu(2008)}]{Harabagiu:AdvancesInOpenDomainQuestionAnswering:2008}
Sanda~M. Harabagiu. 2008.
\newblock Questions and intentions.
\newblock In \emph{Advances in Open Domain Question Answering}.

\bibitem[{Harabagiu et~al.(2000)Harabagiu, Moldovan, Pa{\c{s}}ca, Mihalcea,
  Surdeanu, Bunescu, G{\^\i}rju, Rus, and
  Mor{\u{a}}rescu}]{harabagiu2000falcon}
Sanda~M. Harabagiu, Dan~I. Moldovan, Marius Pa{\c{s}}ca, Rada Mihalcea, Mihai
  Surdeanu, R{\u{a}}zvan Bunescu, Corina~R G{\^\i}rju, Vasile Rus, and Paul
  Mor{\u{a}}rescu. 2000.
\newblock {FALCON: Boosting} knowledge for answer engines.
\newblock In \emph{Proceedings of TREC}.

\bibitem[{Iyyer et~al.(2014{\natexlab{a}})Iyyer, Boyd-Graber, Claudino, Socher,
  and Daum{\'e}~III}]{iyyer2014neural}
Mohit Iyyer, Jordan Boyd-Graber, Leonardo Max~Batista Claudino, Richard Socher,
  and Hal Daum{\'e}~III. 2014{\natexlab{a}}.
\newblock A neural network for factoid question answering over paragraphs.
\newblock In \emph{Proceedings of EMNLP}.

\bibitem[{Iyyer et~al.(2014{\natexlab{b}})Iyyer, Enns, Boyd-Graber, and
  Resnik}]{iyyerpolitical}
Mohit Iyyer, Peter Enns, Jordan Boyd-Graber, and Philip Resnik.
  2014{\natexlab{b}}.
\newblock Political ideology detection using recursive neural networks.
\newblock In \emph{Proceedings of ACL}.

\bibitem[{Jeon et~al.(2005)Jeon, Croft, and
  Lee}]{Jeon:2005:FSS:1076034.1076156}
Jiwoon Jeon, W~Bruce Croft, and Joon~Ho Lee. 2005.
\newblock {Finding Semantically Similar Questions Based on Their Answers}.
\newblock In \emph{Proceedings of SIGIR}.

\bibitem[{Kam(2009)}]{Kam09}
Christopher Kam. 2009.
\newblock \emph{Party Discipline and Parliamentary Politics}.
\newblock Cambridge University Press, Cambridge.

\bibitem[{Kearsley(1976)}]{Kearsley:JournalOfPsycholinguisticResearch:}
Greg~P. Kearsley. 1976.
\newblock {Questions and question asking in verbal discourse: {A}
  cross-disciplinary review}.
\newblock \emph{Journal of Psycholinguistic Research}.

\bibitem[{Lehnert(1978)}]{lehnert1978process}
Wendy~G. Lehnert. 1978.
\newblock \emph{The process of question answering: A computer simulation of
  cognition}.
\newblock Lawrence Erlbaum Associates.

\bibitem[{Louwerse(2012)}]{louwerse2012mechanisms}
Tom Louwerse. 2012.
\newblock Mechanisms of issue congruence: The democratic party mandate.
\newblock \emph{West European Politics}.

\bibitem[{Lytinen and Tomuro(2002)}]{lytinen2002use}
Steven Lytinen and Noriko Tomuro. 2002.
\newblock {The use of question types to match questions in FAQFinder}.
\newblock In \emph{AAAI Spring Symposium on Mining Answers from Texts and
  Knowledge Bases}.

\bibitem[{Mitra and Gilbert(2014)}]{Mitra:ProceedingsOfCscw:2014}
Tanushree Mitra and Eric Gilbert. 2014.
\newblock {The Language that Gets People to Give: Phrases that Predict Success
  on Kickstarter}.
\newblock In \emph{Proceedings of CSCW}.

\bibitem[{Monroe et~al.(2008)Monroe, Colaresi, and
  Quinn}]{Monroe:PoliticalAnalysis:2008}
Burt~L. Monroe, Michael~P. Colaresi, and Kevin~M. Quinn. 2008.
\newblock Fightin' {Words}: {Lexical} feature selection and evaluation for
  identifying the content of political conflict.
\newblock \emph{Political Analysis}.

\bibitem[{Niculae et~al.(2015)Niculae, Suen, Zhang, Danescu-Niculescu-Mizil,
  and Leskovec}]{niculae2015quotus}
Vlad Niculae, Caroline Suen, Justine Zhang, Cristian Danescu-Niculescu-Mizil,
  and Jure Leskovec. 2015.
\newblock {QUOTUS: The} structure of political media coverage as revealed by
  quoting patterns.
\newblock In \emph{Proceedings of WWW}.

\bibitem[{Pedregosa et~al.(2011)Pedregosa, Varoquaux, Gramfort, Michel,
  Thirion, Grisel, Blondel, Prettenhofer, Weiss, Dubourg, Vanderplas, Passos,
  Cournapeau, Brucher, Perrot, and Duchesnay}]{scikit-learn}
F.~Pedregosa, G.~Varoquaux, A.~Gramfort, V.~Michel, B.~Thirion, O.~Grisel,
  M.~Blondel, P.~Prettenhofer, R.~Weiss, V.~Dubourg, J.~Vanderplas, A.~Passos,
  D.~Cournapeau, M.~Brucher, M.~Perrot, and E.~Duchesnay. 2011.
\newblock {scikit-learn}: {Machine} learning in {P}ython.
\newblock \emph{Journal of Machine Learning Research}.

\bibitem[{Pitkin(1967)}]{Pitkin67}
Hanna~F. Pitkin. 1967.
\newblock \emph{The Concept of Representation}.
\newblock University of California Press, Oakland.

\bibitem[{Proksch and Slapin(2011)}]{ProkschSlapin11}
Sven-Oliver Proksch and Jonathan Slapin. 2011.
\newblock Parliamentary questions and oversight in the {European Union}.
\newblock \emph{European Journal of Political Research}.

\bibitem[{Ranganath et~al.(2016)Ranganath, Hu, Tang, Wang, and
  Liu}]{ranganath2016identifying}
Suhas Ranganath, Xia Hu, Jiliang Tang, Suhang Wang, and Huan Liu. 2016.
\newblock Identifying rhetorical questions in social media.
\newblock In \emph{ICWSM}.

\bibitem[{Ravi et~al.(2014)Ravi, Pang, Rastogi, and Kumar}]{Ravi:Icwsm:2014}
Sujith Ravi, Bo~Pang, Vibhor Rastogi, and Ravi Kumar. 2014.
\newblock Great question! {Question} quality in community {Q\&A}.
\newblock In \emph{Proceedings of ICWSM}.

\bibitem[{Rheault et~al.(2016)Rheault, Beelen, Cochrane, and
  Hirst}]{rheault2016measuring}
Ludovic Rheault, Kaspar Beelen, Christopher Cochrane, and Graeme Hirst. 2016.
\newblock Measuring emotion in parliamentary debates with automated textual
  analysis.
\newblock \emph{PloS {One}}.

\bibitem[{Sachdeva and Kumaraguru(2017)}]{Sachdeva:2017:CSC:2998181.2998292}
Niharika Sachdeva and Ponnurangam Kumaraguru. 2017.
\newblock Call for service: {Characterizing} and modeling police response to
  serviceable requests on {Facebook}.
\newblock In \emph{Proceedings of CSCW}.

\bibitem[{Shtok et~al.(2012)Shtok, Dror, Maarek, and
  Szpektor}]{shtok2012learning}
Anna Shtok, Gideon Dror, Yoelle Maarek, and Idan Szpektor. 2012.
\newblock Learning from the past: Answering new questions with past answers.
\newblock In \emph{Proceedings of WWW}.

\bibitem[{Spirling and McLean(2007)}]{SpirlingMcLean07}
Arthur Spirling and Iain McLean. 2007.
\newblock {UK OC OK}? {Interpreting} optimal classification scores for the
  {U.K. House of Commons}.
\newblock \emph{Political Analysis}.

\bibitem[{Tan et~al.(2016)Tan, Niculae, Danescu-Niculescu-Mizil, and
  Lee}]{tan2016winning}
Chenhao Tan, Vlad Niculae, Cristian Danescu-Niculescu-Mizil, and Lillian Lee.
  2016.
\newblock {Winning arguments: Interaction dynamics and persuasion strategies in
  good-faith online discussions}.
\newblock In \emph{Proceedings of WWW}.

\bibitem[{Thomas et~al.(2006)Thomas, Pang, and
  Lee}]{Thomas:2006:GOV:1610075.1610122}
Matt Thomas, Bo~Pang, and Lillian Lee. 2006.
\newblock {Get {Out} the {Vote}: {Determining} support or opposition from
  congressional floor-debate transcripts}.
\newblock In \emph{Proceedings of EMNLP}.

\bibitem[{Treude et~al.(2011)Treude, Barzilay, and
  Storey}]{treude2011programmers}
Christoph Treude, Ohad Barzilay, and Margaret-Anne Storey. 2011.
\newblock How do programmers ask and answer questions on the web?
\newblock In \emph{Proceedings of ICSE}.

\bibitem[{Wang et~al.(2017)Wang, Beauchamp, Shugars, and Qin}]{wangwinning}
Lu~Wang, Nick Beauchamp, Sarah Shugars, and Kechen Qin. 2017.
\newblock Winning on the merits: {The} joint effects of content and style on
  debate outcomes.
\newblock \emph{TACL}.

\bibitem[{Webb and Farrell(1999)}]{webb1999party}
Paul Webb and David~M Farrell. 1999.
\newblock Party members and ideological change.
\newblock \emph{Critical Elections: British Parties and Voters in Long-Term
  Perspective}.

\bibitem[{Zhang et~al.(2016)Zhang, Kumar, Ravi, and
  Danescu-Niculescu-Mizil}]{Zhang:Naacl:2016}
Justine Zhang, Ravi Kumar, Sujith Ravi, and Cristian Danescu-Niculescu-Mizil.
  2016.
\newblock {Conversational flow in Oxford-style debates}.
\newblock In \emph{Proceedings of NAACL}.

\bibitem[{Zhang et~al.(2017)Zhang, Sheng, Lau, and
  Abebe}]{Zhang:2017:DDP:3038912.3052701}
Wei~Emma Zhang, Quan~Z. Sheng, Jey~Han Lau, and Ermyas Abebe. 2017.
\newblock Detecting duplicate posts in programming {QA} communities via latent
  semantics and association rules.
\newblock In \emph{Proceedings of WWW}.

\end{thebibliography}
\bibliographystyle{emnlp_natbib}

\clearpage
\section*{Appendix}
\label{sec:appendix}

\newcommand{\appendixsize}{\footnotesize}
\newcommand{\appendixspace}{\vspace{0.1cm}}
\newcommand{\appendixtype}{\textbf}

\newcommand{\appendixem}[1]{{#1}}
\begin{table*}[t]
\begin{center}
\begin{tabular}{p{0.6cm}  p{14.5cm}  }
\multicolumn{1}{l}{\appendixsize \textbf{Type}} &  \multicolumn{1}{c}{\appendixsize \textbf{Interpretation and examples}}\\
\midrule
\multirow{7}{0.5cm}{ {\rotatebox[origin=r]{90}{\appendixsize \appendixtype{0: Issue update}}}} &	 \appendixsize Requests for information or updates about a current event, issue, or policy.  Typically the policy refers to a genuinely `national' concern, rather than a partisan issue for which the major parties may have differing views.\appendixspace\\
	  &  \appendixsize \appendixem{Q:} \textbf{Will} the Minister also \textbf{update} the House on whether any decisions have been made on the post-2014 UK contribution to Afghanistan?\\
	  &  \appendixsize \appendixem{Q:} \textbf{What} more can the Government \textbf{do} to \textbf{ensure} that each pupil has a single point of contact [for mental health issues] throughout their education?\\
	  &  \appendixsize \appendixem{Q:} \textbf{What} further steps \textbf{can} we \textbf{take} to resolve [the] terrible situation [in Syria]?
	  \appendixspace\\
	  &  {\appendixsize \appendixem{Question motifs:} \{what,are$\leftarrow$taking\}, \{will$\leftarrow$update\}, \{what$\leftarrow$do, do$\rightarrow$ensure\}, \{what$\leftarrow$take, can$\leftarrow$take\}}\\
\midrule
\multirow{9}{0.5cm}{ {\rotatebox[origin=r]{90}{\appendixsize \appendixtype{1: Shared concerns}}}} &	 \appendixsize Straightforward factual question with no strong ideological underpinnings; answers are typically vague and involve explaining that the government takes it seriously, will continue to do so and will consult with the relevant stakeholders.\appendixspace\\
	  &  \appendixsize \appendixem{Q:} \textbf{May} I \textbf{urge} the Minister to concentrate on tough penalties for people who get involved in alcohol-induced antisocial behaviour?\\
	  &  \appendixsize \appendixem{Q:} \textbf{Will} the Secretary of State \textbf{look carefully at} reports that houses built to house the soldiers will block off the rising sun at the summer equinox [at] the Stonehenge?\\
	  &  \appendixsize \appendixem{Q:} \textbf{Will} [my hon. Friend] \textbf{raise} [the matter of] product placement \textbf{with} people in the film industry [when he meets them]? \appendixspace\\
	  &  {\appendixsize \appendixem{Question motifs:} \{will,~take$\rightarrow$*\}, \{may$\leftarrow$urge\}, \{will$\leftarrow$look, look$\rightarrow$at, look$\rightarrow$carefully\}, \{raise$\rightarrow$will, raise$\rightarrow$with\} }\\
\midrule
\multirow{9}{0.5cm}{ {\rotatebox[origin=r]{90}{\appendixsize \appendixtype{2: Narrow factual}}}} &	 \appendixsize Narrow queries about relatively minor policy issues which are either extremely local in nature (perhaps referring to implications for a given constituency) or limited in scope (constituting a small issue within a much broader context of policy).  Questions are ‘precise’ if not especially penetrating.  Answers typically explain the ministry response in narrow, non-ideological terms.\appendixspace\\
	  
	  &  \appendixsize \appendixem{Q:} \textbf{What} funding will be \textbf{given to} the new postgraduate institute at the Edinburgh College of Dentistry?\\
	  &  \appendixsize \appendixem{Q:} \textbf{Will} the Minister \textbf{make} a statement \textbf{on} where we are with the website for the [National Health Service appraisals toolkit for general practitioners]?\\
	  &  \appendixsize \appendixem{Q:} \textbf{What will happen} to the French if they are found guilty of infringing the rules of the Commission? 
	  \appendixspace\\
	  &  {\appendixsize \appendixem{Question motifs:} \{what$\leftarrow$made\}, \{what$\leftarrow$happen, will$\leftarrow$happen\}, \{what$\leftarrow$given, given$\rightarrow$to\}, \{will$\leftarrow$make, make$\rightarrow$on\} }\\
\midrule
\multirow{9}{0.5cm}{ {\rotatebox[origin=r]{90}{\appendixsize \appendixtype{3: Prompt for comment}}}} &	 \appendixsize Requests for comments, or information, especially on a meeting that has taken place between a minister and constituents, colleagues, or opposite numbers, the contents of which would not normally be immediately accessible to MPs.  The asker seeks that the minister clarifies policy where none might presently exist.\appendixspace\\
	  &  \appendixsize \appendixem{Q:} \textbf{Will} the Secretary of State \textbf{tell} us which Minister has been appointed to be responsible for green economic growth? \\
	  &  \appendixsize \appendixem{Q:} \textbf{Can} [the Secretary of State] \textbf{confirm} whether train platform capacity will \textbf{be} a part of the discussions between the city council and Network Rail?\\
	  &  \appendixsize \appendixem{Q:} \textbf{What would} the Prime Minister \textbf{say} to a borough council which is considering rejecting Government funding and instead [taxing] my constituents more?
	  \appendixspace\\
	  &  {\appendixsize \appendixem{Question motifs:} \{what$\leftarrow$say,say$\rightarrow$to\}, \{will$\leftarrow$tell\}, \{can$\leftarrow$confirm, confirm$\rightarrow$be\}, \{what$\leftarrow$say, would$\leftarrow$say\}}\\

\end{tabular}
\end{center}
 \caption{Interpretation for the first four types, with examples of representative questions and motifs.}
 \label{tab:appendix1}
 \end{table*}

\begin{table*}[t]
\begin{center}
\begin{tabular}{p{0.6cm}  p{14.5cm}  }
\multicolumn{1}{l}{\appendixsize \textbf{Type}} &  \multicolumn{1}{c}{\appendixsize \textbf{Interpretation and examples}}\\
\midrule
\multirow{8}{0.5cm}{ {\rotatebox[origin=r]{90}{\appendixsize \appendixtype{4: Agreement}}}} &	 \appendixsize Airing a laudatory remark about a policy that the minister and MP clearly already agree on.  Often these questions effectively serve as attempts to curry favor with the minister and bolster their (mutual) party.\appendixspace\\
	  &  \appendixsize \appendixem{Q:} \textbf{Is} it not \textbf{important} that the Department continues its excellent work [in] building flood defeneces? \\
	  &  \appendixsize \appendixem{Q:} \textbf{Does} [the Secretary of State] \textbf{agree with} me that part of protecting Britain's national interests is that Britain should develop relationships with emerging economies?\\
	  &  \appendixsize \appendixem{Q:} \textbf{Does} the Minister \textbf{agree} that UK taxpayers \textbf{need} to be considered at every single step of the way when it comes to our aid spending?
	  \appendixspace\\
	  &  {\appendixsize \appendixem{Question motifs:} \{does$\leftarrow$agree, agree$\rightarrow$is\}, \{is$\rightarrow$important\}, \{does$\leftarrow$agree, agree$\rightarrow$with\}, \{does$\leftarrow$agree, agree$\rightarrow$need\}}\\
\midrule
\multirow{8}{0.5cm}{ {\rotatebox[origin=r]{90}{\appendixsize \appendixtype{5: Self-promotion} }}} &	 \appendixsize Here, ``awareness'' is entirely rhetorical: either the minister is aware and agrees or the minister is not aware and will investigate.  These questions allow the question asker to be seen to be bringing local concerns to broader attention, but in a way that is more assertive than in type 2 (\textbf{narrow factual}).
\appendixspace\\
	 	&  \appendixsize \appendixem{Q:} \textbf{Has} the right hon. Gentleman \textbf{considered} compulsory postal voting , and moving polling day from a Thursday to the weekend? \\
	  &  \appendixsize \appendixem{Q:} In considering the role of local tribunals , \textbf{will} my hon Friend \textbf{take} account of the recommendation \textbf{in} the Oglesby report?\\
	  &  \appendixsize \appendixem{Q:} \textbf{Will} the Minister \textbf{reconsider} his reply to my [colleague], [in light of] the situation [on] school leavers applying for technical positions?
	  \appendixspace\\
	  &  {\appendixsize \appendixem{Question motifs:} \{is$\rightarrow$aware\}, \{considered$\leftarrow$has\}, \{will$\leftarrow$take, take$\rightarrow$in\}, \{will$\leftarrow$reconsider\} }\\
\midrule
\multirow{7}{0.5cm}{ {\rotatebox[origin=r]{90}{\appendixsize \appendixtype{6: Concede, accept} }}} &	 \appendixsize Aggressive demand for minister to concede to, or accept, a fault.  The premise of such questions is that the minister has been incompetent, or that the government has the wrong policy; these questions do not constitute a genuine attempt to obtain information.
\appendixspace\\
	  &  \appendixsize \appendixem{Q:} \textbf{Is} it \textbf{not} now completely \textbf{true} that the Labour Government are out of touch with gut British instincts? \\
	  &  \appendixsize \appendixem{Q:} \textbf{Will} [the Secretary] \textbf{acknowledge} the importance of not completely abandoning the research on sustainable biofuels?\\
	  &  \appendixsize \appendixem{Q:} \textbf{Will} [the Deputy Prime Minister] now \textbf{concede} to the House that the Royal Mail was sold off too cheaply?\appendixspace\\
			  &  {\appendixsize \appendixem{Question motifs:} \{will$\leftarrow$accept\}, \{is$\rightarrow$not,~is$\rightarrow$true\}, \{will$\leftarrow$acknowledge\}, \{will$\leftarrow$concede\}}\\
\midrule
\multirow{8}{0.5cm}{ {\rotatebox[origin=r]{90}{\appendixsize \appendixtype{7: Condemnatory}  }}} &	 \appendixsize Similar to type 6 (\textbf{concede, accept}) but more aggressive and hectoring in tone, asking the minister to explain themselves and be contrite on the basis of very broad ideological premises that are difficult to answer without `self-incrimination'.  These questions often rely on rhetorical grandstanding and lacks any subtlety or policy detail on which minister can comment precisely.\appendixspace\\
  &  \appendixsize \appendixem{Q:} When members of the armed forces are facing a pay freeze, \textbf{how can} the Secretary \textbf{justify} bonuses to senior offices in the civil service? \\
	  &  \appendixsize \appendixem{Q:} \textbf{Why does} the right hon Gentleman not have an industrial strategy to build that recovery?\\
	  &  \appendixsize \appendixem{Q:} \textbf{Will} the Government now \textbf{apologise} for their complacent decision to scrap the future jobs fund, [given that] long-term youth unemployment is rising?\appendixspace\\
			  &  {\appendixsize \appendixem{Question motifs:} \{can$\leftarrow$explain\}, \{how$\leftarrow$justify, can$\leftarrow$justify\}, \{why does\}, \{will$\leftarrow$apologise\}}\\

\end{tabular}
\end{center}
 \caption{Interpretation for the last four types, with examples of representative questions and motifs.}
  \label{tab:appendix2}
 \end{table*}

\subsection*{A.1 \ \ \ Further examples of question types}
Tables \ref{tab:appendix1} and \ref{tab:appendix2} provide further examples of representative questios and motifs from each of the eight question types we induce on our dataset of parliamentary question periods. Additionally, we include extended interpretations of each of these types, provided by the second author, a political scientist with domain expertise in the UK parliamentary setting. %

\subsection*{A.2 \ \ \ Details about data filtering decisions}
Here we provide further details about how we selected the subset of 50,152 questions which was used in the analyses described in Sections \ref{sec:validation} \&~\ref{sec:insights}. First, we restrict our analysis to the questions which consist of only one question, as opposed to a series of questions (as delimited by multiple sentences ending in question marks; constituting 52\% of the data). We omit these multi-question utterances in order to ensure that we don't confound effects arising from different co-occurring question types. Next, we only include questions for which information about the asker and answerers' party affiliations and ministership positions are available (such data is provided consistently from the Blair government onwards). Finally, we omit questions asked by opposition members who are specifically appointed by their party to {\em shadow} a government minister, and are hence obliged by their appointment to ask more critical questions. Our choice to omit such questions eliminates the possibility that our observed differences in question preference are driven by official appointment.

\subsection*{A.3 \ \ \ Details about the labeled PMQ dataset}

Here we provide further details about the size of the labeled Questions to the Prime Minister (PMQ) dataset from \citet{Batesetal2014} used in the quantitative validation of our typology (Section \ref{sec:validation}). The dataset contains 931 standard, 186 helpful and 139 unanswerable questions. Additionally, 445 answers to questions are labeled as \textit{answered}, while 305 are labeled as \textit{not answered}; the rest are labeled as \textit{deferred answer}, meaning the Prime Minister did not have the knowledge or capability to provide an answer. 

We restrict all analysis done using the unanswered vs.~answered labels to \textit{standard} questions, i.e., questions for which the PM has an opportunity to provide a legitimate answer.  

The \textit{helpful vs.~standard} classification task contains 264 train and 108 test examples; the \textit{unanswerable vs.~standard} classification task contains 190 train and 86 test examples; the \textit{unanswered vs. answered} classification task contains 166 train and 84 test examples.

\subsection*{A.4 \ \ \ Example motif subgraph}

\begin{figure*}
\centering
\includegraphics[width=0.9\textwidth]{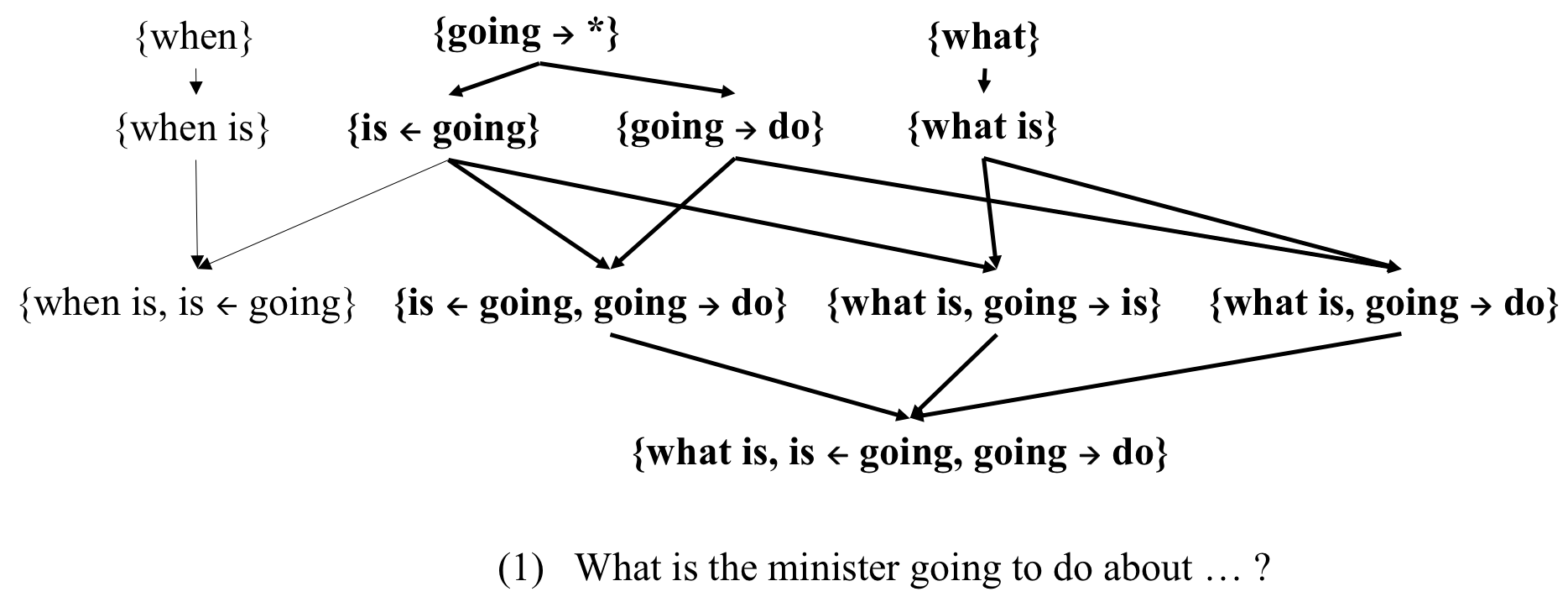}
\caption{A section of the motif DAG $\mathcal{M}$ and the subgraph $\mathcal{M}_q$ (in bold) representing the phrasing of the question in example (1).  For clarity, redundant and irrelevant nodes and edges are not shown.}
\label{fig:subgraph}
\end{figure*}

Figure \ref{fig:subgraph} illustrates a section of the motif DAG $\mathcal{M}$ and highlights (in bold) the subgraph $\mathcal{M}_q$ corresponding to the phrasing of the question in example (1).  Nodes that are higher in the graph serve as general representations, and capture similarities between broad sets of phrasings: e.g., \{is$\leftarrow$going\\{going$\rightarrow$do, is$\leftarrow$going, what is\}} groups together example (1) with a question like ``When is he going to get a grip on [the scandal]''.  By projecting such motifs into our latent question-answer space (Section \ref{sec:qtypes}) we capture characteristics shared between these phrasings and allow for generalizability.  Nodes which are lower in the graph constitute more specific representations, disambiguating between phrasings.  In particular, sink motifs serve as the most specific representation of a question, delineating the region of the latent space (and thus the question type) that best captures its phrasing. For instance, this additional specificity allows us to draw contrasts between ``\textbf{What is} the minister \textbf{going} to \textbf{do} [about the policy]?'' (sink motif: \{what is, is$\leftarrow$going, going$\rightarrow$do\}) and the more aggressive ``\textbf{When is} he \textbf{going} to get a grip on [the scandal]?'' (sink motif: \{when is, is$\leftarrow$going\}).

\newpage

\end{document}